\documentclass{IEEEtran}

\usepackage{subfigure}
\usepackage{cite}
\usepackage{amsmath,amssymb,amsfonts}
\usepackage[ruled,linesnumbered]{algorithm2e}

\usepackage{color}
\usepackage{graphicx}
\usepackage{textcomp}
\usepackage{booktabs}
\usepackage{multirow}
\usepackage{bbding}
\usepackage{bm}

\def\BibTeX{{\rm B\kern-.05em{\sc i\kern-.025em b}\kern-.08em
	T\kern-.1667em\lower.7ex\hbox{E}\kern-.125emX}}

\begin{document}

\author{
	\vskip 1em
	{
Lei Ren, \IEEEmembership{Member, IEEE,}
Haiteng Wang, \IEEEmembership{Student Member, IEEE,}
Yuanjun Laili, \IEEEmembership{Member, IEEE}
	} 
\thanks{
    {	
This work has been submitted to the IEEE for possible publication. Copyright may be transferred without notice, after which this version may no longer be accessible
    
        Lei Ren, Haiteng Wang, and Yuanjun Laili are with School of Automation Science and Electrical Engineering, Beihang University, Beijing 100191, China. (email: renlei@buaa.edu.cn, wanghaiteng@buaa.edu.cn, lailiyuanjun@buaa.edu.cn).
    }
}
}
\markboth{IEEE TRANSACTIONS ON XXXX, VOL. XX, NO. XX, XX 2024}%
{Shell \MakeLowercase{\textit{et al.}}: Bare Demo of IEEEtran.cls for IEEE Journals}

\title{Diff-MTS: Temporal-Augmented Conditional Diffusion-based AIGC for Industrial Time Series Towards the Large Model Era}
\maketitle
\begin{abstract}

Industrial Multivariate Time Series (MTS) is a critical view of the industrial field for people to understand the state of machines. However, due to data collection difficulty and privacy concerns, available data for building industrial intelligence and industrial large  models is far from sufficient. Therefore, industrial time series data generation is of great importance. Existing research usually applies Generative Adversarial Networks (GANs) to generate MTS.  However, GANs suffer from unstable training process due to the joint training of the generator and discriminator.
This paper proposes a temporal-augmented conditional adaptive diffusion model,  termed Diff-MTS, for MTS generation. It aims to better handle the complex temporal dependencies and dynamics of MTS data. Specifically, a conditional Adaptive Maximum-Mean Discrepancy (Ada-MMD) method has been proposed for the controlled generation of MTS, which does not require a classifier to control the generation. It improves the condition consistency of the diffusion model. Moreover, a Temporal Decomposition Reconstruction UNet (TDR-UNet) is established to capture complex temporal patterns and further improve the quality of the synthetic time series. Comprehensive experiments on the C-MAPSS and FEMTO datasets demonstrate that the proposed Diff-MTS performs substantially better in terms of diversity, fidelity, and utility compared with GAN-based methods. These results show that Diff-MTS facilitates the generation of industrial data, contributing to intelligent maintenance and the construction of industrial large  models.
	
\end{abstract}


\section{Introduction}
\label{sec:introduction}
Industrial Multivariate Time Series (MTS) plays a vital role in anomaly detection\cite{li2023filter}, remaining useful life prediction\cite{ren2023dlformer,qin2022slow} and fault diagnosis\cite{ren2023single}. It is the basis for training deep learning models with high inference accuracy.  However, multivariate time series data exhibit heterogeneity due to different time scales, noise levels, and different latent characteristics. Coupled with the data privacy concerns of different industrial companies, available industrial time series data for specific scenarios is rare. Data shortage has become a severe problem that hinders intelligent maintenance research and the construction of industrial large models.

Data generation techniques provide a means of alleviating data shortage and enhancing the test environment. The commonly applied AI-Generated Content (AIGC) methods for generating time series data include variational autoencoders (VAEs)\cite{CHEN2021vae,dai2023variational,zhao2022new} and generative adversarial networks (GANs)\cite{Vu2023Generative,behera2021generative,zhang2022multiclass}, etc. For example, Chen et al.\cite{CHEN2021vae} utilize VAE to generate trajectories to address the trajectory generation problem, which employs an LSTM network to learn the characteristics of trajectories. A modified Wasserstein auto-encoder (MWAE) is presented in the article\cite{zhao2022new} to generate highly similar fault data, which introduces the sliced Wasserstein distance for measuring distributional differences. However, the VAE usually fails to generate realistic samples due to its reconstruction loss function. Another recent promising method to generate samples is the GAN-based model, which employs a discriminator and generator through an adversarial process to create realistic data. CVAE-GAN\cite{zhang2022multiclass} is a conditional variational generative adversarial network for bearing fault diagnosis, which adopts a classifier to distinguish different classes of fault data. CPI-GAN\cite{xiong2023controlled} integrates the physical information to the generative adversarial network to generate synthetic degradation trajectories, which improves the accuracy of downstream tasks. These methods reduce the time and effort required for manual data collection and improve the accuracy of the model. Although generative models have achieved success in some generation tasks, there are still some challenges in MTS data generation.

Firstly, GANs-based models are prone to non-convergence and unstable training process\cite{li2023reducing, saxena2021generative,yin2020wasserstein}. Specifically,  MTS data is of low quality with high sampling frequency and strong noises, which makes it difficult for the generator in the GAN to learn the patterns in the time series data. Meanwhile, the discriminator can easily distinguish between real and synthetic samples. The adversarial nature between the generator and discriminator demands substantial effort to stabilize the training process.

Secondly, MTS data have different conditions (e.g., fault classes, health indicators), and it's challenging for the generative model to generate data with specific conditions (consistency between input condition information and generated data). To guide the synthetic samples, some methods\cite{behera2021generative,bao2017cvae,zhang2022multiclass} leverage discriminator or classifier to achieve controlled generation of MTS data. However, these methods require joint training additional network structure, which reduces the condition consistency of controlled generation. 

Thirdly, MTS data are usually non-smooth and involve complex temporal dependencies, leading to difficulties in generating MTS data. That is, time series at the current moment are correlated with time series at previous moments, and this relationship is non-stationary and difficult to predict. This causes challenges for the generative model to extract trend information, thus reducing the similarity between the synthetic data and original data.

Denoising Diffusion Probabilistic Models (DDPM)\cite{ho2020denoising} is proposed to model the data distribution by simulating a diffusion process that evolves the input data towards a target distribution. Although some works have achieved excellent results in generating images with continuous pixel values\cite{yang2023diffmic, nichol2021improved, dhariwal2021diffusion}, they lack sufficient MTS generation capacity and still struggle to address the above issues. To bridge the research gap, a temporal-augmented conditional adaptive model, namely Diff-MTS, is proposed for the synthesis of industrial multivariate time series. It addresses the non-convergence and unstable training issues in previous works and alleviates the diffusion model's weak ability to generate sensor time series. Specifically, our contribution can be summarized as follows:

1) A temporal-augmented diffusion model is proposed for MTS generation, which incorporates conditional adaptation and temporal decomposition reconstruction, addressing the deficiency of traditional diffusion models in generating MTS with complex temporal dependencies.

2) A classifier-free, conditional Adaptive Maximum-Mean Discrepancy (Ada-MMD) diffusion method is proposed for the controlled generation of MTS, addressing the limitation of jointly training classifiers to control the generation. The adaptive-MMD mechanism adaptively learns mutual information in the latent space, which enhances the condition consistency of the diffusion model.

3) A Temporal Decomposition Reconstruction UNet (TDR-UNet) is proposed to denoise and recover MTS data. The TDR mechanism is proposed to extract the underlying patterns and trend information of sensor time series, which enhances diffusion model's ability to generate high-fidelity MTS data.

4) A variety of experiments are employed to comprehensively measure the diversity, fidelity, and usefulness of the generated MTS on multiple datasets. The results verify the outstanding performance of the Diff-MTS and the great potential of diffusion models for generating MTS data.

The rest of this article is organized as follows. Section II discusses the related works about industrial multivariate data synthesis and diffusion models. Section III detailly describes the framework of the proposed method. Besides, experimental results and analysis are presented in Section IV. Finally, Section V concludes this paper.

\section{Related Works}
\subsection{Industrial Multivariate Time Series Synthesis}
Previous studies have investigated the application of generative networks for synthesizing multivariate time series data. One commonly used approach is the variational autoencoders (VAEs) \cite{kingma2013auto,pan2022imputation,zhao2022new}, which is used to generate new samples that are similar to the original data For instance, Chen et al. \cite{CHEN2021vae} use a VAE to generate trajectories to tackle the trajectory generation problem, employing an LSTM network to learn the trajectory characteristics. 
The literature\cite{zhao2022new} presents a modified Wasserstein auto-encoder (MWAE) to generate highly similar fault data. This method introduces the sliced Wasserstein distance for measuring distributional differences. In contrast, GAN-based models\cite{goodfellow2014generative, ducoffe2019anomaly, behera2021generative, zhang2022multiclass} employ a discriminator and generator through an adversarial process to create realistic data. Lu et al. \cite{lu2021deep} introduce a GAN architecture that combines the AE and LSTM network to monitor the RUL of bearings. CPI-GAN \cite{xiong2023controlled} integrates physical information into the GAN to generate synthetic degradation trajectories, which can enhance the accuracy of downstream tasks. 

Unlike GAN-based or VAE-based models requiring additional architecture for training (e.g., the discriminator in GAN or the encoder in VAE), we adopt the diffusion model for MTS generation task. This model use a forward diffusion process and doesn't require addition network in training, avoiding mode collapse and training instability issues from the joint training of two networks.

\subsection{Diffusion Models}
Diffusion models have recently been proposed to model data distributions using forward and reverse processes. The forward process gradually injects noise into real-world data, whereas the reverse process generates realistic data by removing noise. The fundamental research on diffusion models is proposed and theoretically supported by Sohl-Dickstein et al.\cite{sohl2015deep}.   Two significant advances in diffusion models, i.e., Denoising Diffusion Probabilistic Models (DDPM)\cite{ho2020denoising} and conditional diffusion models on image synthesis\cite{dhariwal2021diffusion}, have demonstrated remarkable abilities in image generation\cite{croitoru2023diffusion}. These advances have catalyzed further developments and applications of diffusion models in various domains. Recently, diffusion models have been successfully extended to generate audio and ECG time series \cite{kong2020DiffWave} \cite{adib2023synthetic}. DiffWave \cite{kong2020DiffWave} involves a raw audio synthesis technique based on a diffusion probabilistic model, which has achieved significant results in audio quality.

While diffusion models are commonly applied to image tasks, our research explores their potential in MTS data generation. Specifically, we expand the traditional diffusion model to a conditional adaptive generative model within the sensor time series and develop a temporal decomposition reconstruction module for MTS data.

\section{Temporal-augmented Conditional Adaptive Diffusion Model for ITMS Generation}
In this section, we present a temporal-augmented conditional adaptive diffusion model for MTS data generation. The model comprises the Denoised Diffusion Probabilistic Model, Conditional Ada-MMD Diffusion Method, and Temporal Decomposition Reconstruction UNet (TDR-UNet). Each of them will be introduced in detail as follows. 
\subsection{Denoised Diffusion Probabilistic Model}
\begin{figure}[!tpb]
	\centerline{\includegraphics[width=1\columnwidth]{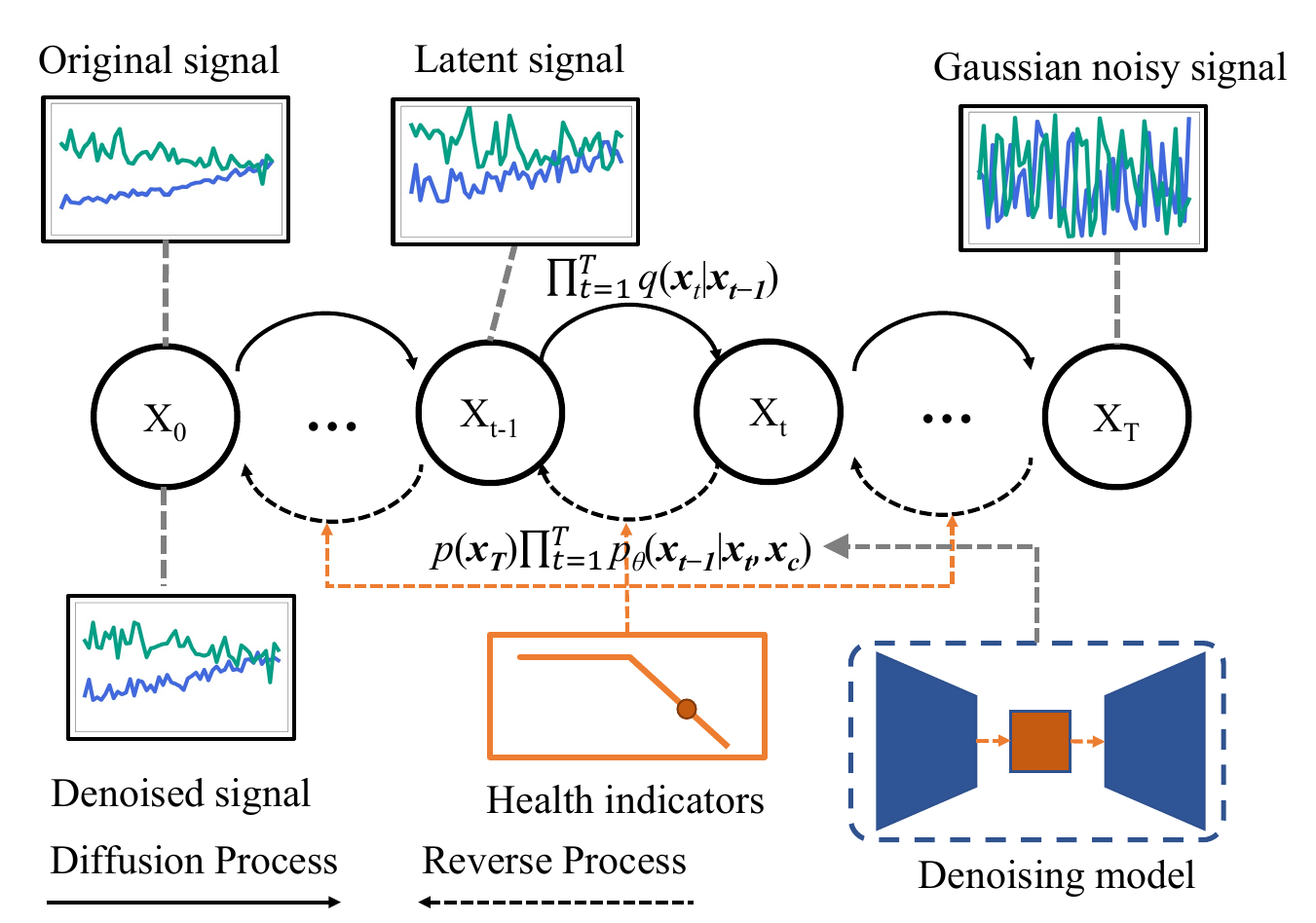}}
	\caption{Illustration for diffusion process and reverse process. During diffusion process, noise is gradually added to the original signal to become Gaussian noisy signal. In the reverse process, the noisy signal is recovered to the original signal by estimating the added noise.}
	\label{fig:diffusion Framework}
\end{figure}
Compared to the VAE-based methods, diffusion models possess high-dimensional latent variables through a diffusion process, enabling them to generate high-quality samples. Additionally, diffusion models employ a fixed diffusion learning procedure that avoids the instability training problem of GAN-based methods. 
In essence, our approach extends the principles established by DDPM\cite{ho2020denoising} to the domain of one-dimensional multivariate signal generation. 

This diffusion model consists of two main stages, as shown in Fig. \ref{fig:diffusion Framework}: a diffusion process and a reverse process.
In the diffusion process, Gaussian noise is added to the original time series $\boldsymbol{x_0}$. This noise is dependent on the time step $t$, which is sampled from a uniform distribution within the range of $[1, T]$. We refer to these noisy variables as $\boldsymbol{x_1, ..., x_t, .., x_T}$. The process of diffusion can be defined by a Markov chain which remains fixed, starting from the data \(\boldsymbol{x_0}\) and ending with the latent variable \(\boldsymbol{x_T}\):

\begin{equation}
    \label{eq:diffusion}
    q(\boldsymbol{x_{0:T} | x_0}) = \prod_{t=1}^T q(\boldsymbol{x_t | x_{t-1}})
\end{equation}
where \(q(\boldsymbol{x_t | x_{t-1}}) := \mathcal N(\boldsymbol{x_t}; \sqrt{1 - \beta_t}\boldsymbol{x_{t-1}}, \beta_t \boldsymbol{I})\), \(\beta_t\) is the noise schedule.  $T$ and $t$ are diffusion total timestep  and current timestep, respectively. This entire process gradually transforms data \(\boldsymbol{x_0}\) into Gaussian distribution variables \(\boldsymbol{x_T}\) when $T \to \infty $.

We employ the reparameterization technique to modify the diffusion process in Eq.  (\ref{eq:diffusion}) to enhance its efficiency. The value of $x_t$ at any random timestep $t$ can be computed as follows:
\begin{equation}
    \label{eq:reparameter}
     \boldsymbol{x_t} = \sqrt{\bar{\alpha}_t} \boldsymbol{x_0} + \sqrt{1 - \alpha_t} \boldsymbol{\epsilon}
\end{equation}
where $\boldsymbol{\epsilon} \sim \mathcal{N}(0, \boldsymbol{I})$ is Gaussian noise; $\bar{\alpha}_t = \prod_{i=1}^t \alpha_i$; and $\alpha_t = 1 - \beta_t$.

The reverse process takes on the role of denoising $\boldsymbol{x_t}$ to recover $\boldsymbol{x_0}$ so that it can eventually recover data samples from the Gaussian noise. To achieve this, the noise added during the diffusion process is predicted through a trained denoising model.
The reverse step is similarly defined by a Markov chain, as in the diffusion step.
\begin{equation}
    p_{\theta}(\boldsymbol{x_{0:T-1} | x_T}) = \prod_{t=1}^T p_{\theta}(\boldsymbol{x_{t-1} | x_t})
\end{equation}
where \(\theta\)  are the parameters of the denoising model.

\begin{figure*}[!tpb]
	\centerline{\includegraphics[width=2.1\columnwidth]{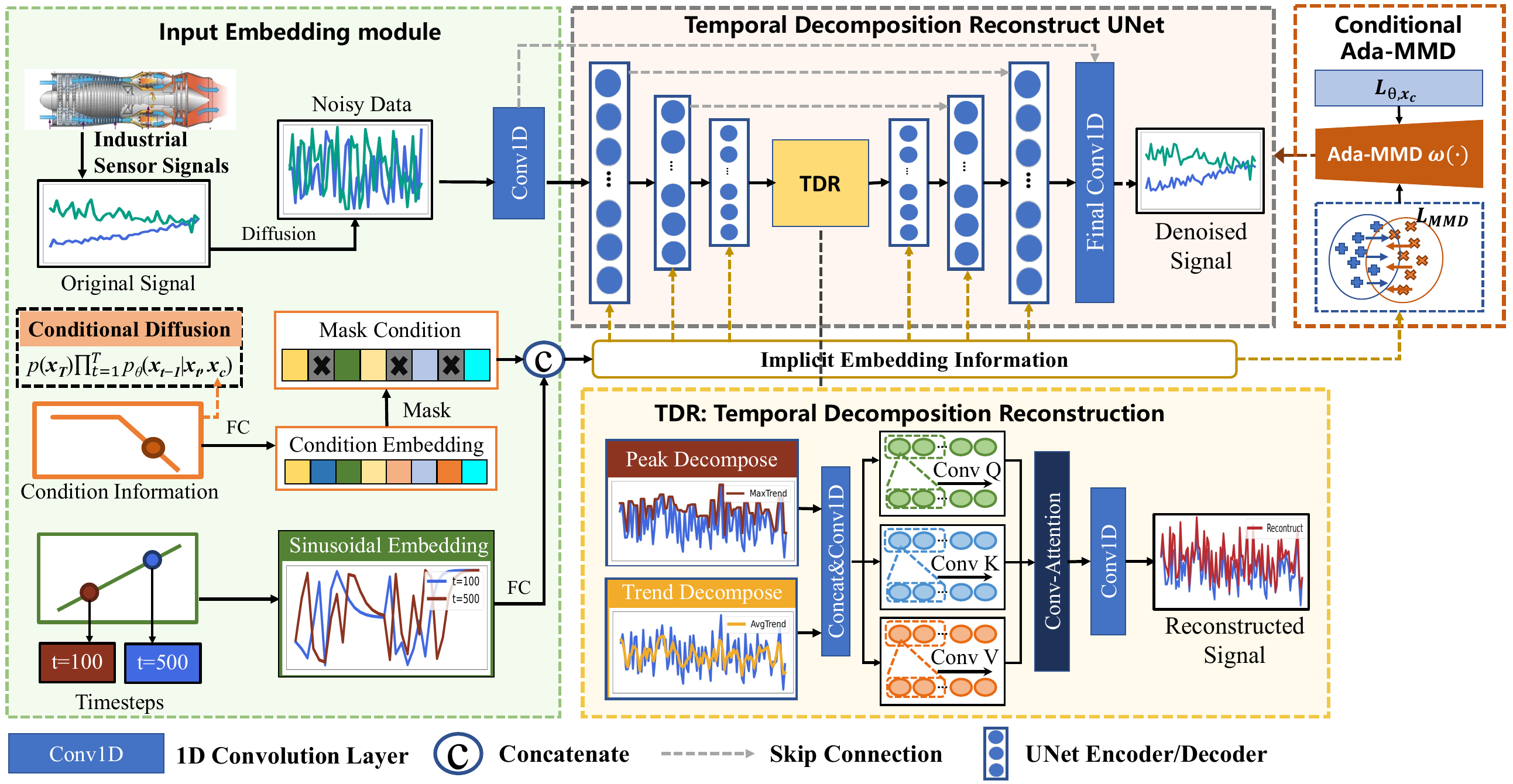}}
	\caption{Overview of the Temporal Decomposition Reconstruction UNet Model.}
	\label{fig:framework}
\end{figure*}

In the original DDPM work, a linear noise schedule was used. Nonetheless, in complex signal processing scenarios, this schedule failed to produce optimal results. The strong linear schedule led to rapid degradation of the noisy time series into pure noise, resulting in faster information loss during the noise introduction phase. To mitigate this problem, we implemented a cosine schedule as suggested in the paper\cite{nichol2021improved}. 
\begin{equation}
\label{eq:linear schedule}
f(t) = \frac{1}{T-t+1} 
\end{equation}

\begin{equation}
\label{eq:cosine schedule}
f(t) = \cos\left(\frac{t}{T} + \frac{s}{1+s} \cdot \frac{\pi}{2}\right)^2
\end{equation}
where Eq. (\ref{eq:linear schedule}) is the linear schedule function. Eq. (\ref{eq:cosine schedule}) is the cosine schedule function. $s$ is an offset parameter set to $s$ = 0.008. The cosine schedule is designed to preserve information within the noisy time series for a more extended period during the noising process steps.

\subsection{Conditional Ada-MMD Diffusion Method}
To achieve the synthesis of multivariate time series under specific conditions, it is necessary to incorporate condition information, such as equipment health indicators, into the diffusion model. Therefore, we develop a conditional diffusion model that does not require an explicit classifier and considers the health indicator of the equipment as the condition. Then, to ensure the condition consistency of the diffusion model, a Adaptive Maximum-Mean Discrepancy (Ada-MMD) regularization loss is introduced to adaptively capture mutual information. This facilitates the alignment of the generated samples under specific conditions with the original samples, promoting their similarity. 

First, we add the condition information \(x_c\) to the reverse process. The conditional form of reverse process can be formulated as follows:
\begin{equation}
\begin{aligned} 
\label{eq:classift}
p_{\theta}(\boldsymbol{x_{0:T} | x_c}) &= p(x_T) \prod_{t=1}^T p_{\theta}(\boldsymbol{x_{t-1} | x_t, x_c}) \\
p_{\theta}(\boldsymbol{x_{t-1} | x_t, x_c}) :&= \boldsymbol{\mathcal{N}} (\boldsymbol{x_{t-1}; \mu_{\theta}(x_t}, t|\boldsymbol{x_c}), \boldsymbol{\sigma_{\theta}(x_t}, t|\boldsymbol{x_c) I})
\end{aligned}
\end{equation}

Then, the parameterizations of $\mu_\theta$ and $\sigma_{\theta}$ can be defined by:
\begin{equation}
\label{eq:mean}
\boldsymbol{\mu_\theta(x_t}, t | \boldsymbol{x_c}) = \frac{1}{\sqrt{\alpha_t}} \left( \boldsymbol{x_t} - \frac{\beta_t}{\sqrt{1 - \alpha_t}} \boldsymbol{\epsilon_\theta(x_t}, t | \boldsymbol{x_c}) \right)
\end{equation}
\begin{equation}
\label{eq:var}
\boldsymbol{\sigma_{\theta}(x_t}, t|\boldsymbol{x_c}) = \frac{1 - \bar{\alpha}_{t-1}}{1 - \bar{\alpha}_t} \beta_t
\end{equation}
where $\epsilon_\theta$ denotes the estimated noise. $\beta_t$ is the noise schedule. $ \bar{\alpha}_t$ has been defined in the Eq. (\ref{eq:reparameter}). According to the Eqs. (\ref{eq:classift})-(\ref{eq:mean}), the time series  can be recovered by the estimated noise. 

Similar to the standard reverse process, the original signal can be deduced from the noise distribution if it is predicted.
Afterward, we input the noisy variable $x_t$, timestep $t$, and condition information $x_c$ into our denoising model UNet to estimate the noise distribution.
To incorporate the condition into the denoising model, condition information is included in the optimization objective. In the conditional reverse process, we aim to minimize the noise estimation loss by following:
\begin{equation}
L_{\theta,x_c} = \mathbb{E}_{x_0 \sim D, \epsilon \sim N(0,I), t} \left\| \boldsymbol{\epsilon - \epsilon_\theta(x_t}, t|\boldsymbol{x_c}) \right\|_2^2 
\end{equation}
where $D$ represents the data distribution. This function can correspond to the noise estimation errors. Given a pair of samples, after training using the DDPM approach, the sampled model contains the guidance of \(\epsilon\) in a way that directs the noise towards the direction of $\theta$, enabling noise reduction. 

To enhance the similarity between real and synthetic time series, our study introduces the conditional Ada-MMD regularization loss. Specifically, the noisy variable $x_{t}$ is derived from the diffusion process at time step $t$ and an Ada-MMD regularization loss with conditional information is then computed to reduce the divergence between the Gaussian distribution and the sampled noise distribution. 

\begin{equation}
\begin{aligned}
L_{\text{MMD}}(n,m) &= K(n, n') - 2K(m, n) + K(m, m') \\
& with\ n = \boldsymbol{\epsilon}, m = \boldsymbol{\epsilon_\theta(x_t}, t|x_c)
\end{aligned}
\end{equation}
\begin{equation}
\label{eq:condition LOSS}
L_{\rm diff} = (1-\omega)L_{\theta,x_c} +  \omega L_{\text{MMD}}(n,m)
\end{equation}
 where $\boldsymbol{\epsilon}$ denotes real noise, $\boldsymbol{\epsilon_\theta}$ are the predicted noise by the denoising model with condition information. $K(\cdot)$ represents a positive definite kernel designed to reproduce distributions in the high-feature dimension space. While the general noise estimation loss $L_{\theta,x_c}$ captures information from the European distance, the conditional MMD regularization maps the data to a high-dimensional feature space to capture similarities between distributions. The parameter $\omega$ functions as an adaptable coefficient, calibrating the significance of the MMD regularization loss within the overall objective function. This configuration permits dynamic adjustment of the model's focus on capturing mutual information, which is based on the learning progress and data characteristics.

The classifier-free guidance approach guides the diffusion model's training without relying on explicit classifier-based control time series. This method introduces a conditional Ada-MMD loss to reduce the difference between distributions, effectively contributing to the optimization process.

\subsection{Temporal Decomposition Reconstruction UNet}
\label{sec:UNet}
A Temporal Decomposition Reconstruction UNet Model (TDR-UNet) is designed to tackle the task of generation. The model mainly includes two components, the input embedding module and the Temporal Decomposition Reconstruction UNet module. Specifically, the model embeds the inputs, including the latent variable $\boldsymbol{x_t}$, the condition $x_c$ , and the timestep $t$. The UNet encoders and decoders\cite{ho2020denoising} are used for predicting the noise that is added to the original data during the diffusion process. To capture the temporal dependency of the MTS data, a decomposition reconstruction mechanism is executed between the encoder and decoder. The detailed architecture is shown in Fig. \ref{fig:framework}.

\subsubsection{Input Embedding Module}
To preserve the temporal relationships within the data, we initiate the embedding transformation for each input sample.  First, to facilitate the input to be better learned by the subsequent UNet, the latent variable $\boldsymbol{x_t}$ is embedded using a 1D convolution layer.
\begin{equation}
\boldsymbol{x_{t_{emb}}} = {\rm Conv1D}(\boldsymbol{x_t})
\end{equation}
where $\text{Conv1D}(\cdot)$ represents 1D  convolution layer. $\boldsymbol{x_{t_{emb}}}$ denotes the embeded time series. 

Next, the discrete timestep $t$ is embedded sinusoidally with a two-layer fully connected (FC) network to a continuous feature $\boldsymbol{t_{\rm emb}}$, enabling the network to understand data over time. 

\begin{equation}
\begin{aligned}
t_{\rm pos} &= {\rm PosEmbed}(t) \\
\boldsymbol{t_{\rm emb}} &= {\rm FC(GeLU(FC(}t_{\rm pos})))
\end{aligned}
\end{equation}
where $\rm PosEmbed(\cdot)$ denotes sinusoidal position embedding methods\cite{vaswani2017attention}. $t_{pos}$ refers to the initial positional embedding of the timestep $t$. GeLU is an activation function.

 In order to enable the input signal to contain the target information, we transformed condition information into a vector via an FC network. 
\begin{equation}
\boldsymbol{x_{c_{emb}}} = {\rm Mask}_\alpha({\rm FC}(x_c))
\end{equation}
where \(\text{FC}(\cdot)\) denotes the fully-connected layer.  In contrast to the one-hot processing for discrete features, we employ an FC network to embed continuous conditions in high-dimension space. Mask($\cdot$) is a mask encoding function. Specifically, we set a parameter $ \alpha $ to adjust the degree of condition information. For the embedded condition vectors, $\alpha$ of them will be set to random values. For example, the larger the value of $\alpha$, the larger the random value in the embedded condition vector will be. This will result in less conditional information.

\subsubsection{Temporal Decomposition Reconstruction UNet Module}
\begin{figure}[!tpb]
	\centerline{\includegraphics[width=1\columnwidth]{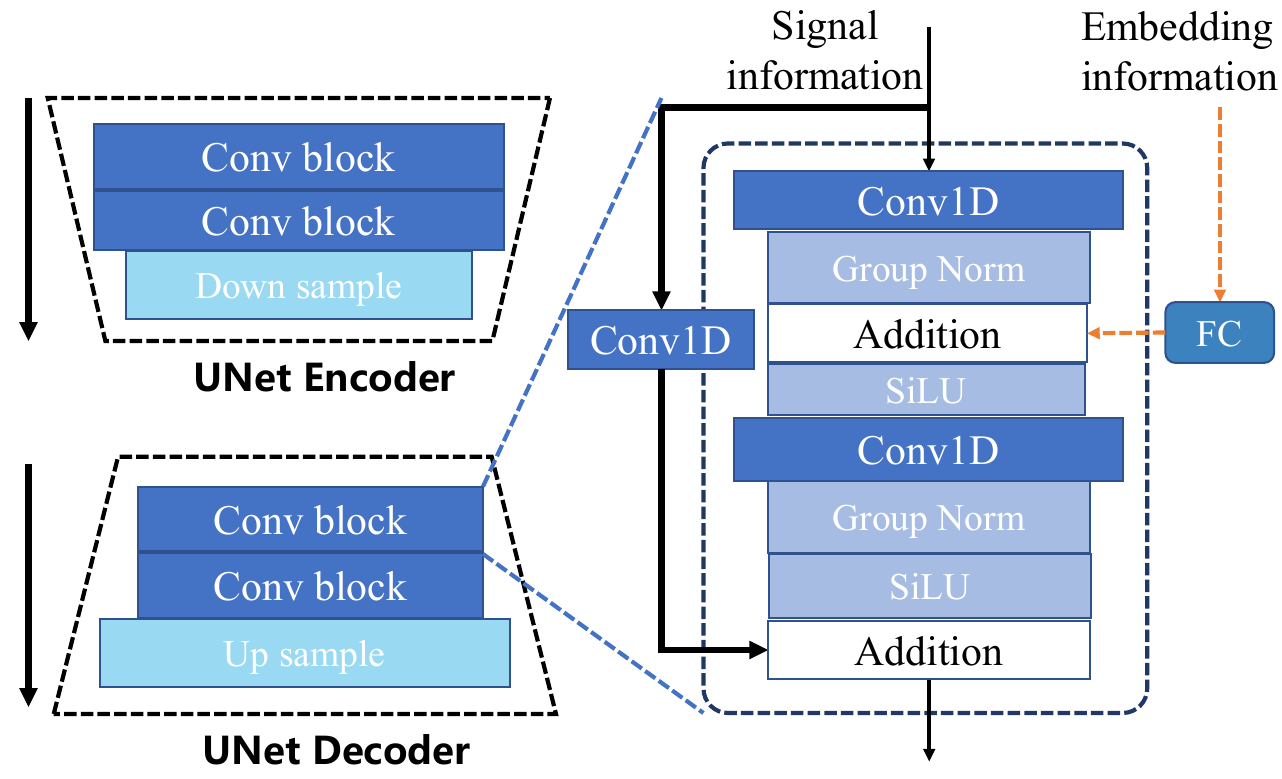}}
	\caption{Structure of the UNet encoder and decoder.}
	\label{fig:convblock}
\end{figure}

In the proposed UNet method, the network architecture is constructed based on the U-Net framework [15], consisting primarily of three encoders and three decoders. Each encoder features two sequential 1D convolution blocks, followed by a downsampling operation. Within each convolution block, two convolutional layers are employed: the first layer conducts a convolution operation on the input signal, which is then processed by a GroupNorm transformation and activated via a SiLU function. It is illustrated in Fig. \ref{fig:convblock}.

The architecture integrates residual connections that add the input signal information directly to the output in the encoder and after convolution in the decoder. Additionally, condition and timestep information are fed into the first convolution block of each encoder. The decoder consists of two convolution blocks that progressively restore the signal to its original dimensions through convolution operations.  These convolutional layers incorporate skip connections, which fuse features from corresponding layers in the encoder with those in the decoder.

Then, a temporal decomposition reconstruction layer is performed after the UNet encoder. To learn the complex temporal patterns in time series generation context, we apply the technique of time series decomposition on multivariate time series data. Given the input $\boldsymbol{X}$,  average-pooling and max-pooling are employed for the decomposition of $\boldsymbol{X}$, yielding two distinct sets of features: $\boldsymbol{ X_{\rm Trend}}$  and $\boldsymbol{X_{\rm Peak}}$. The trend part assists in capturing the inherent average trend within the data, while the utilization of peak parts serves to accentuate maximal variation. Each represents one of the underlying categories of patterns that are more predictable. Subsequently, these two features are concatenated and fed into a convolution block to integrate the time series information. This decomposition procedure can be described as:

\begin{equation}
\begin{aligned}
   \boldsymbol{ X_{\rm Trend}} &= {\rm AvgPool(Padding}(\boldsymbol{X} )) \\
    \boldsymbol{X_{\rm Peak}} &= {\rm MaxPool(Padding}(\boldsymbol{X} )) \\
    \boldsymbol{F_{\rm dec}} &= {\rm Conv1d(Concat}(\boldsymbol{X_{\rm Trend},X_{\rm Peak}}))  
\end{aligned}
\end{equation}

After time series decomposition, a convolutional attention structure is performed to reconstruct the multivariate time series. The time series is first processed through three 1D convolutional layers to generate separated features. Then, an attention mechanism is executed as follows:

\begin{equation}
\begin{aligned}
\boldsymbol{Q_{\rm conv}} = {\rm Conv}&(\boldsymbol{F_{\rm dec}});   \boldsymbol{K_{\rm conv}} = {\rm Conv}(\boldsymbol{F_{\rm dec}}); \\ 
\boldsymbol{V_{\rm conv}} &= {\rm Conv}(\boldsymbol{F_{\rm dec}}) \\
{\rm Attention}\left(\boldsymbol{Q,K,V}\right) &={\rm softmax}\left(\frac{\boldsymbol{Q_{\rm conv}K^T_{\rm conv}}}{\sqrt{d_{k}}}\right)\boldsymbol{V_{\rm conv}}
\end{aligned}
\end{equation}
where  $\boldsymbol{Q_{\rm conv},K_{\rm conv},V_{\rm conv}} $ are parameter metrics. $\sqrt{d_{k}} $ is the scaling factor. Since the attention layer is capable of highlighting the significant fields that are beneficial for feature representation, this layer facilitates learning the fine-grained information of the average part and max part. Therefore, the network is better to learn the underlying patterns. Subsequently, the model reconstructs the time series information through a one-dimensional convolutional layer.  Finally, the reconstructed information is processed by the UNet decoder layer and a convolution layer to recover the original size.

\subsection{Training and Sampling Procedure}
The proposed methodology begins with constructing a neural network model for noise estimation  of MTS data. The model parameters $\theta$ are then trained by minimizing the loss function calculated by Eq. (\ref{eq:condition LOSS}) through the Adam algorithm. This training procedure, outlined in Algorithm \ref{alg:training}, enhances both training efficiency and convergence speed. The algorithm starts by initializing the model parameter and setting up the training environment. First, Gaussian noise is generated and a diffusion timestep is selected. Then, the corresponding target condition from the condition dataset guides the diffusion process. The latent variable is calculated, followed by an estimation of the added noise based on the model parameters. Next, the model's loss is calculated using a combination of standard noise estimation and Ada-MMD loss, weighted by a learnable factor. After processing all instances in the batch, the model parameters are updated using the Adam optimization algorithm, based on the gradients of the loss. This process is repeated through all epochs until the fully trained model parameters are output, ready for deployment or further validation.

Once the training process is completed, the method generates MTS data for specific operating conditions in an iterative manner. This procedure is presented in Algorithm \ref{alg:sample}. The first step is to determine the total number of diffusion steps and acquire the noise schedules. Next, a random noise sample $x_T$ is sampled from a Gaussian distribution. The typical HI of industrial equipment associated with the data synthesis is selected as the conditional input $x_c$. Along with the random noise and time step, this input is utilized in the model explained in Section \ref{sec:UNet}. The following steps include calculating the noisy data $x_t$ based on the model's output and repeating this procedure until $t = 0$. At this point, the denoised data $x_0$ serves as the generated time series.

\begin{algorithm}[!thpb]

\caption{Conditional Diffusion Models Training}
\label{alg:training}
\small
\KwIn{The model parameter $\theta$, noise schedules $\{\beta_t\}_{t=1}^{T}$, total diffusion  timestep $T$,  industrial monitoring dataset $D$, condition dataset $D_c$, batch size $B$, total epoch number $E$, Adam parameters $\alpha_{\text{Adam}}$, $\beta_{\text{Adam1}}$, $\beta_{\text{Adam2}}$, learnable factor $\omega$}
\KwOut{The complete trained model}
\For{$e = 1, \ldots, E$}{
    \For{$i = 1, \ldots, B$}{
        Sample original signal $x_0 \sim D$, Gaussian noise $\epsilon \sim \mathcal{N}(0, I)$, diffusion timestep $t \sim \text{Uniform}(1, T)$\;
        Choose the corresponding target condition $x_c$ from $Dc$\;
        Calculate latent variliable $x_t \leftarrow \sqrt{\bar{\alpha}_t x_0} + \sqrt{(1 - \bar{\alpha}_t)} \epsilon$\;
        Estimated the added noise $\epsilon_\theta(x_t, t|x_c)$\;
        Calculate loss of the denoising model  $L^{i} \leftarrow (1-\omega)L_{\theta,x_c} +  \omega L_{\text{MMD}}(n,m)$    \;
    }
    Update model's parameters based on Adam algorithm $\theta \leftarrow \text{Adam}(\nabla_\theta \frac{1}{B} \sum_{i=1}^{B} L^{i},  \beta_{\text{Adam1}}, \beta_{\text{Adam2}}, \alpha_{\text{Adam}})$\;
}
\Return{$\theta$}
\end{algorithm}

\begin{algorithm}[!thpb]
\label{alg:sample}
\small
\caption{Conditional Diffusion Models Sampling}

\KwIn{The well-trained model $\theta$, diffusion total timestep $T$, noise schedules $\{ \beta_t \}_{t=1}^T$, condition dataset $D_c$}
\KwOut{Industrial synthetic multivariate time series}
    Sample $x_T \sim N(0, I)$\;
    Obtain $x_c$ from condition dataset $D_c$\;
    \For{$t = T, \ldots, 1$}{
        Sample $\epsilon \sim N(0, I)$\;
        \If{$t = 0$}{
            $\epsilon \leftarrow 0$\;
        }
        Calculate latent variable $x_{t-1}$ according to Eqs.~(\ref{eq:classift})--(\ref{eq:var}):
        $x_{t-1} \leftarrow \frac{1}{\sqrt{\alpha_t}} ( x_t - \frac{\beta_t}{\sqrt{1 - \bar\alpha_t}}   \epsilon_\theta(x_t, t | xc) ) + \frac{1 - \bar\alpha_{t-1}}{1 - \bar\alpha_t} \beta_t \epsilon$\;
    }
    \Return $x_0$\;
    
\end{algorithm}

\section{Experiment}

\subsection{Experimental Data and Setup}
\subsubsection{Dataset}
The C-MAPSS dataset, which characterizes the degradation process of turbofan engines, was released by NASA in 2008 and has since gained widespread use. The dataset comprises data from 21 multi-sensors, including time, pressure, and speed, to record the engine's condition. Not all multi-sensor data is pertinent to equipment HI as multi-sensor data points 1, 5, 6, 10, 16, 18, and 19 have consistent values. Accordingly, we have excluded these data points, which leaves us with 14 multi-sensors\cite{ren2023time}. Additionally, the C-MAPSS dataset comprises four sub-datasets. In particular, sub-datasets FD001 and FD003 include a single operating condition and one fault type. The FD002 subset comprises six operating conditions and one faulty condition, while the FD004 subset comprises two faulty conditions and six operating conditions. These two datasets are more complex and provide a more representative reflection of complex system conditions.

The FEMTO dataset collects data of bearing wear using a rotational speed multi-sensor and a force multi-sensor with a sampling frequency of 25.6 kHz. The dataset represents real-world conditions of bearings under accelerated degradation in three different conditions. In this paper, we have chosen bearings 1-1 and 1-2 as our training dataset for generation experiments under condition 1, with a rotational speed of 1800 RPM and a force of 4000 N. The preprocessing procedures and the bearing end life are also consistent with prior research\cite{ren2022lightweight}.

\begin{center}
\begin{table}[tpb]
		\centering
		\caption{Key hyperparameters of the proposed method }
		\label{tab:parameters}       
\begin{tabular}{cll}
\hline
Module                               & Layer                                & Parameter               \\ \hline
\multirow{5}{*}{Input Embedding}     & Noisy embed.                      & conv(7,32,1)            \\ \cline{2-3} 
                                     & \multirow{2}{*}{Condition embed.} & hidden units = 128      \\
                                     &                                      & output units = 128      \\ \cline{2-3} 
                                     & \multirow{2}{*}{Timestep embed.} & hidden units = 128      \\
                                     &                                      & output units = 128      \\ \hline
\multirow{4}{*}{UNet Encoder}        & \multirow{2}{*}{Conv Nets 1}         & conv(3,32,1)            \\
                                     &                                      & FC: output units = 64    \\ \cline{2-3}
                                     & Conv Nets 2                          & conv(3,32,1)            \\
                                     & Downsample                           & conv(4,32/64/64,2)      \\ \hline
\multirow{3}{*}{Decomposition Layer} & AvgPool                              & kernel\_size=3, stride=1 \\
                                     & MaxPool                              & kernel\_size=3, stride=1 \\
                                     & Conv1D                               & conv(1,64,1)            \\ \hline
\multirow{2}{*}{Reconstruction Layer}   & Conv1D Q,K,V                           & conv(1,384,1)           \\
                                     & Conv1D                               & conv(1,64,1)            \\ \hline
\multirow{6}{*}{UNet Decoder}        & \multirow{2}{*}{Conv Nets 1}         & conv(4,32/64/64,2)      \\
                                     &                                      & FC: output units = 128   \\ \cline{2-3} 
                                     & Conv Nets 2                          & conv(4,32/64/64,2)      \\
                                     & Res Conv1D                           & conv(1,64,1)            \\  \cline{2-3}
                                     & \multirow{2}{*}{Upsample}            & scale\_factor = 2       \\
                                     &                                      & filters = 64/32/32      \\ \hline
Output                               & Conv1D Layer                         & conv(1,14,1)            \\ \hline
\end{tabular}
\end{table}
\end{center}

\begin{table*}[htpb]
	\centering
	\caption{Comparison Results With Other Methods on C-MAPSS Datasets. Bold represents the best results, underlining represents the second best results.}
	\label{tab:comparation-CMAPSS} 
\begin{tabular}{lccccccc}
\hline
Methods  & TTS-GAN\cite{li2022tts} & TimeGAN\cite{yoon2019time} & GAN-LSTM\cite{lu2021deep} & SSSD\cite{lopezalcaraz2022diffusionbased}   & Tabddpm\cite{kotelnikov2023tabddpm}        & DiffWave\cite{kong2020DiffWave} & \textbf{Diff-MTS}       \\ \hline
\multicolumn{8}{c}{Discriminative Score (Lower the better)}                                    \\ \hline
FD001-24 & 0.998   & 0.964   & 0.951    & 0.868  & \underline{0.805}          & 0.828    & \textbf{0.640}  \\
FD001-48 & 0.995   & 0.919   & 0.942    & 0.911  & \textbf{0.889} & 0.914    & \underline{0.897}           \\
FD001-96 & 0.987   & 0.935   & 0.953    & 0.950  & \underline{0.811}          & 0.904    & \textbf{0.611}  \\
FD002-24 & 0.994   & 0.957   & 0.951    & 0.843  & \textbf{0.802} & 0.857    & \underline{0.808}           \\
FD002-48 & 0.991   & 0.936   & 0.944    & 0.731  & 0.796          & \underline{0.676}    & \textbf{0.664}  \\
FD002-96 & 0.998   & 0.941   & 0.949    & \underline{0.785}  & 0.813          & 0.801    & \textbf{0.755}  \\
FD003-24 & 0.998   & 0.920   & 0.953    & 0.805  & \underline{0.792}          & 0.841    & \textbf{0.706}  \\
FD003-48 & 0.998   & 0.983   & 0.955    & 0.917  & \underline{0.898}          & 0.931    & \textbf{0.886}  \\
FD003-96 & 0.983   & 0.951   & 0.947    & 0.970  & \underline{0.805}          & 0.925    & \textbf{0.672}  \\
FD004-24 & 0.989   & 0.917   & 0.959    & 0.862  & \underline{0.807}          & 0.843    & \textbf{0.695}  \\
FD004-48 & 0.991   & 0.965   & 0.958    & \underline{0.719}  & 0.809          & 0.784    & \textbf{0.659}  \\
FD004-96 & 0.998   & 0.946   & 0.945    & 0.818  & \underline{0.800}          & 0.829    & \textbf{0.722}  \\ \hline
\multicolumn{8}{c}{Predictive Score (Lower the better)}                                        \\ \hline
FD001-24 & 85.581  & 70.536  & 33.223   & \underline{19.864} & 28.359         & 20.431   & \textbf{17.547} \\
FD001-48 & 94.107  & 41.497  & 75.488   & 23.100 & 30.636         & \underline{18.085}   & \textbf{13.616} \\
FD001-96 & 85.291  & 37.933  & 65.259   & \underline{17.364} & 36.088         & 21.774   & \textbf{14.472} \\
FD002-24 & 87.154  & 41.230  & 90.656   & 27.724 & 30.922         & \underline{26.950}   & \textbf{24.039} \\
FD002-48 & 95.273  & 46.780  & 55.138   & \underline{25.431} & 29.837         & 27.184   & \textbf{24.408} \\
FD002-96 & 87.211  & 41.132  & 45.194   & \underline{27.628} & 32.956         & 30.122   & \textbf{26.152} \\
FD003-24 & 72.454  & 85.800  & 25.705   & \underline{21.208} & 25.165         & 21.552   & \textbf{20.377} \\
FD003-48 & 86.811  & 53.928  & 59.376   & 25.959 & 26.274         & \underline{23.749}   & \textbf{17.184} \\
FD003-96 & 72.354  & 85.170  & 34.037   & \underline{22.946} & 28.994         & 28.433   & \textbf{15.554} \\
FD004-24 & 86.216  & 51.086  & 58.782   & 43.531 & 34.557         & \underline{34.382}   & \textbf{29.247} \\
FD004-48 & 94.756  & 86.609  & 45.654   & 39.764 & 33.604         & \underline{32.793}   & \textbf{27.072} \\
FD004-96 & 84.803  & 50.805  & 48.106   & 41.186 & 42.764         & \underline{32.515}   & \textbf{23.167} \\ \hline
\end{tabular}
\end{table*}

\begin{table}[htpb]
\centering
\caption{Comparison Results With Other Methods on FEMTO Datasets}
\label{tab:comparation-FEMTO} 
\begin{tabular}{lcccccc}
\hline
Dataset--Metric & \multicolumn{3}{c}{FEMTO--Predictive.}  & \multicolumn{3}{c}{FEMTO--Discriminative.} \\ \hline
Seq length      & 80             & 160            & 320            & 80              & 160            & 320            \\ \hline
TTS-GAN\cite{li2022tts}          & 0.862          & 0.975          & 0.942          & 1.000               & 0.989          & 0.991          \\
TimeGAN\cite{yoon2019time}         & 0.802          & 0.882          & 0.889          & 0.99            & 0.979          & 0.982          \\
GAN-LSTM\cite{lu2021deep}        & 0.728          & 0.829          & 0.927          & 0.986           & 0.998          & 0.983          \\
DiffWave\cite{kong2020DiffWave}        & 0.213          & 0.204          & 0.239          & 0.864           & 0.741          & 0.741          \\
Diff-MTS         & \textbf{0.143} & \textbf{0.148} & \textbf{0.158} & \textbf{0.683}  & \textbf{0.683} & \textbf{0.663} \\ \hline
\end{tabular}
\end{table}

\subsubsection{Evaluation Metrics}
To comprehensively evaluate the performance of the generative model, we use a variety of metrics\cite{yoon2019time,xiong2023controlled} to assess the quality of the generated data, focusing on three key criteria: diversity, fidelity, and usefulness. Diversity: Synthetic data should be widely distributed to cover the range of original data. Fidelity: Synthetic data should be indistinguishable from the original data. Usefulness: Synthetic data should be useful when used for the same predictive purposes, such as training on synthetic data and testing on original data.

\textbf{\textit{Visualization}}: we conducted t-SNE and PCA analyses on both the synthetic and original datasets, providing a 2-dimensional representation for qualitatively assessing the diversity of the generated samples. 

\textbf{\textit{Discriminative Score}}: To evaluate fidelity, we employ a two-layer LSTM to train a classification model for time series. We classify each original sequence as "real" and each generated sequence as "not real", and then train a 2-layer LSTM classifier to discriminate between these two classes. We divide the dataset comprising of synthetic and original time series into 70\% for the training set, 10\% for the validation set, and 20\% for the test set. The classification accuracy of the test set provides a numerical measure of the similarity between the two datasets.

\textbf{\textit{Predictive Score}}: To evaluate usefulness, we utilize the synthetic dataset to train a time series prediction model using a two-layer LSTM. This model predicts the HI of equipment over each input sequence. We then evaluate the trained model's performance on the original test dataset using root mean absolute error (RMSE) as the metric. 

By utilizing these methods, we conduct a thorough assessment of the quality of the generated data in terms of its diversity, fidelity, and usefulness.

\subsubsection{Implementation Details}
Table \ref{tab:parameters} lists the key hyperparameters of the proposed model. To simplify the table, ``conv(7,32,1)" means the 1D convolution layer with kernel sizes of 7, filters of 32, and strides of 1. Moreover, ``conv(4,32/64/64,2)" of the downsample layer means that there are three 1D convolution layers with kernel sizes of 7 and strides of 2. The filters are 32, 64, and 64, respectively. This simplified expression also applies to other convolutional layers. During the training, we used the Adam optimizer for 70 epochs. For training the predictive model and discriminative model, we set the epochs to 40. 
Each experiment was repeated five times to reduce the randomness, and the results were average values. 

\begin{figure*}[htpb]
	\centering
	{
		\includegraphics[width=0.235\linewidth]{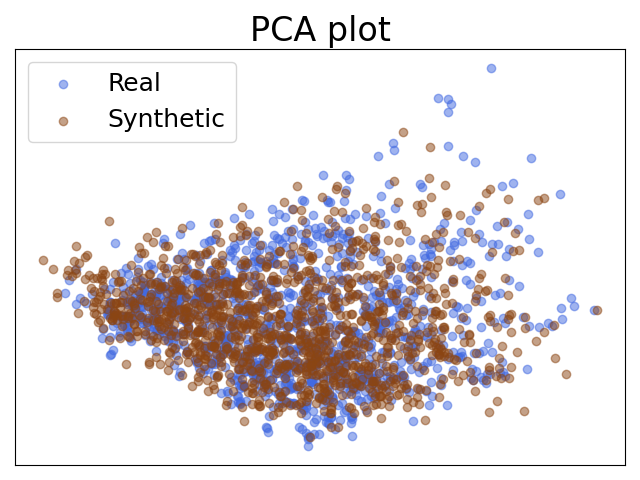}}
	\hspace{0in}    
	{
		\includegraphics[width=0.235\linewidth]{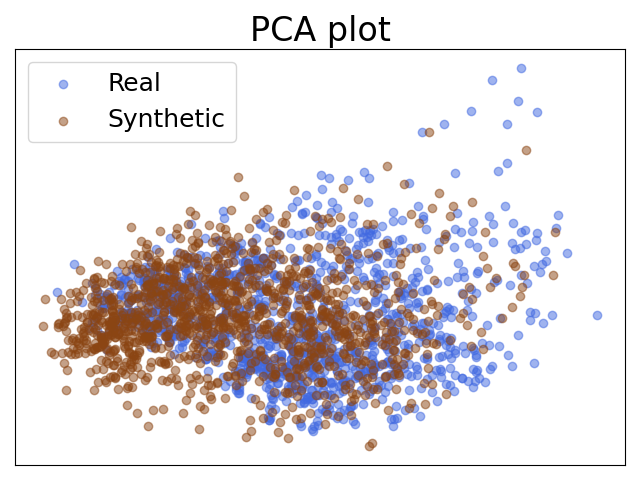}}
	\hspace{0in}    
	{
		\includegraphics[width=0.235\linewidth]{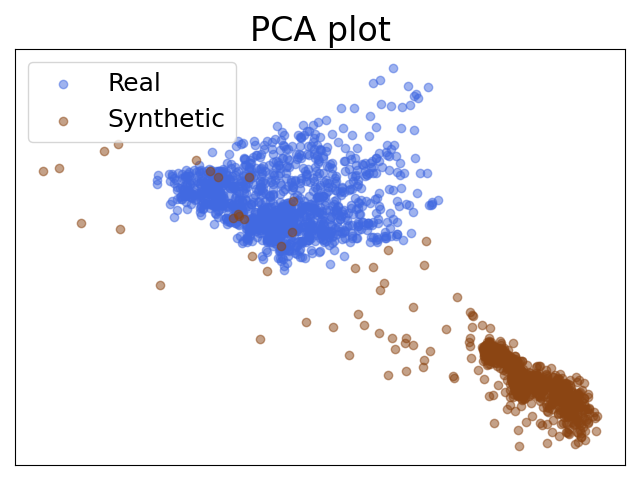}}
	\hspace{0in}
	{
		\includegraphics[width=0.235\linewidth]{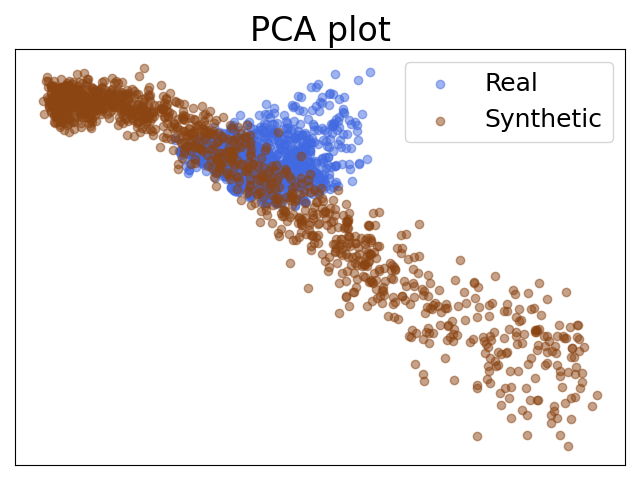}}
  
        \subfigure[Diff-MTS]{
            \includegraphics[width=0.233\linewidth]{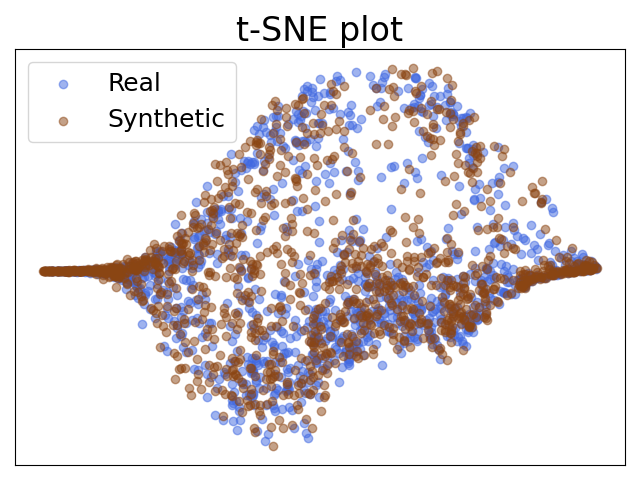}}
        \hspace{0in}  
        \subfigure[DiffWave]{
            \includegraphics[width=0.233\linewidth]{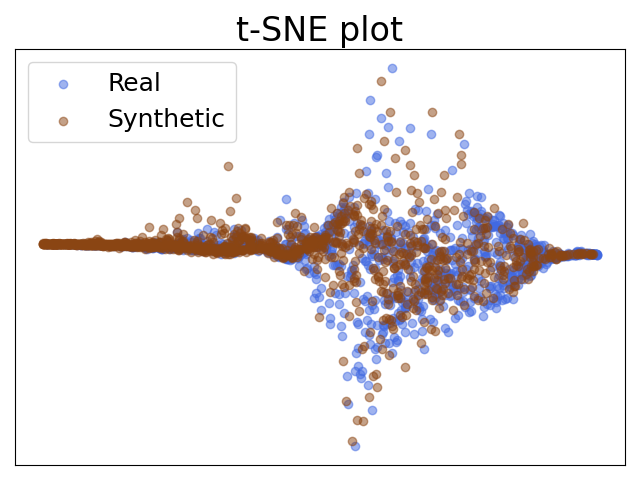}}
        \hspace{0in}  
        \subfigure[TimeGAN]{
            \includegraphics[width=0.233\linewidth]{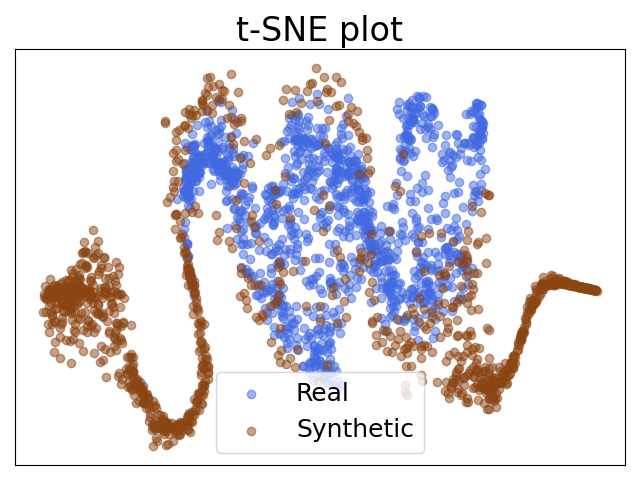}}
        \hspace{0in}  
        \subfigure[TTS-GAN]{
            \includegraphics[width=0.233\linewidth]{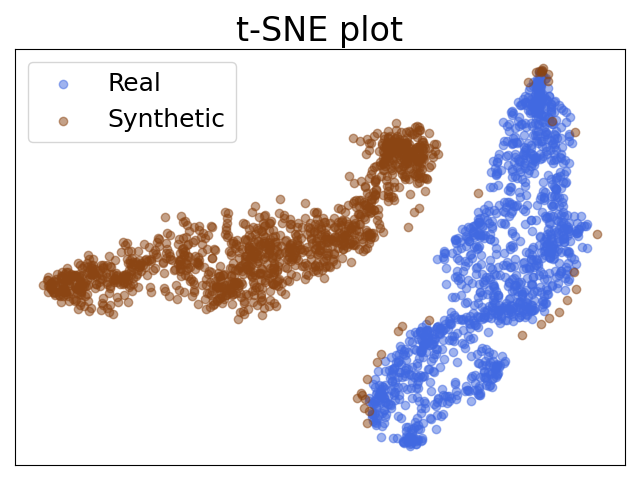}}
        \hspace{0in}  
	\caption{ PCA visualization  (1$^{\rm st}$ row) and t-SNE visualization (2$^{\rm nd}$ row) on FD001 dataset. Each column presents the two visualizations of the four methods. Blue represents the original samples, and brown represents the synthesis samples. Two points overlapping more means that the two data types are more similar.}
	\label{fig:prediction plot1}
\end{figure*}

\begin{figure*}[htpb]
	\centering
	\subfigure[Original data]{
		\includegraphics[width=\linewidth]{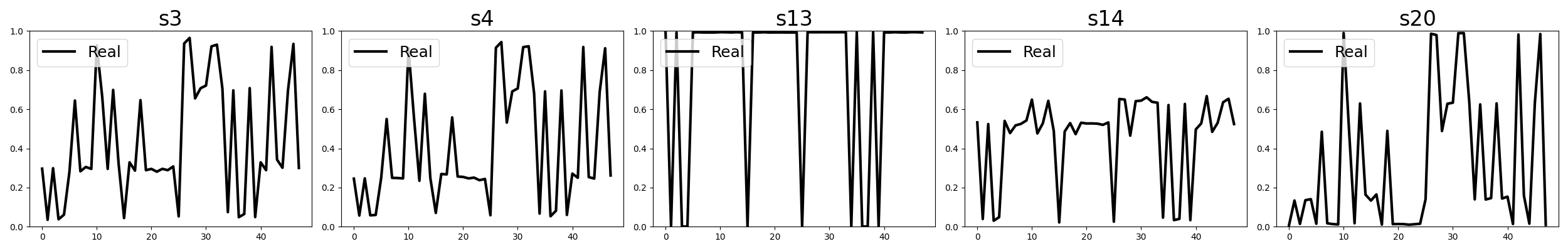}}
	\hspace{0in}    
	\subfigure[Diff-MTS]{
		\includegraphics[width=\linewidth]{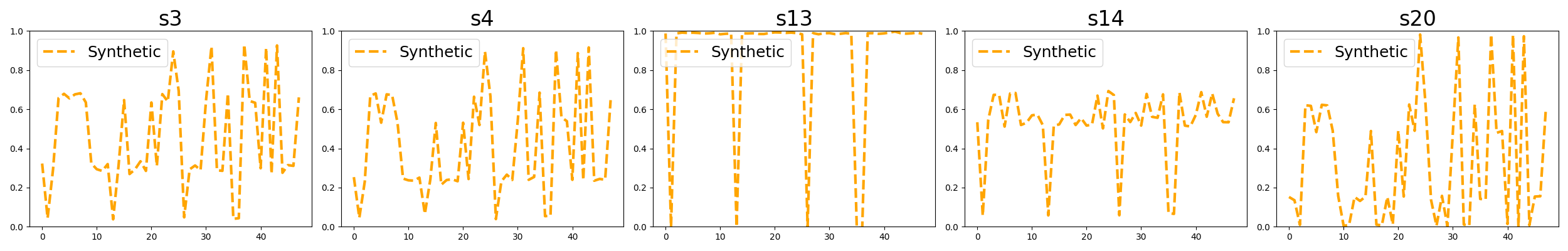}}
	\hspace{0in}    
	\subfigure[DiffWave]{
		\includegraphics[width=\linewidth]{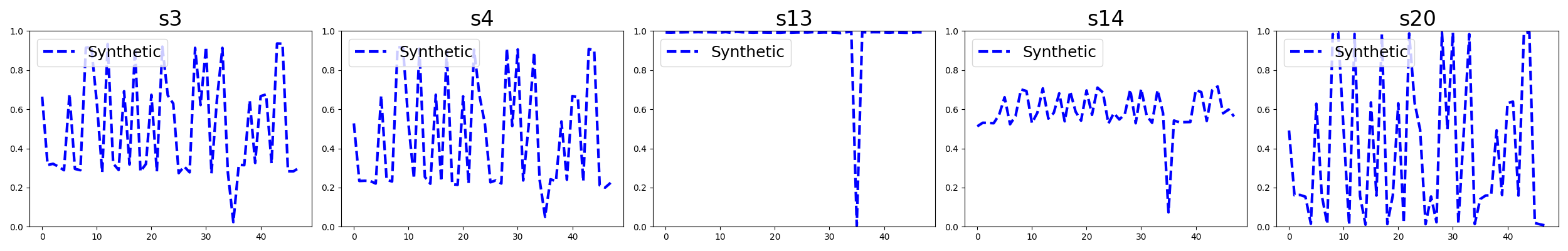}}
	\hspace{0in}    
	\subfigure[TimeGAN]{
		\includegraphics[width=\linewidth]{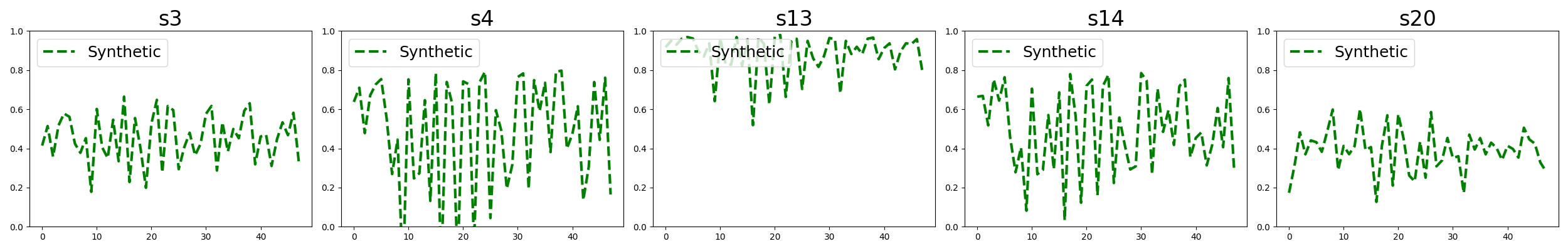}}
	\caption{Several visualizations of prediction results. Black curves represent the original time series. The gold, blue and green curves represent the synthetic time series of Diff-MTS, DiffWave and TimeGAN, respectively. }
	\label{fig:curves}
\end{figure*}

\begin{table*}[htpb]
\centering
\caption{Ablation study of TDR-UNet. }
\label{tab:ablation-att} 
\begin{tabular}{cccccccccccccc}
\hline
Dataset       & \multicolumn{3}{c|}{FD001}                                                                             & \multicolumn{3}{c|}{FD002}                                                                                      & \multicolumn{3}{c|}{FD003}                                                                    & \multicolumn{3}{c}{FD004}                                                                              & \multirow{2}{*}{Average} \\ \cline{1-13}
Length        & 24                         & 48                                  & \multicolumn{1}{c|}{96}             & 24                                  & 48                                  & \multicolumn{1}{c|}{96}             & 24                         & 48                                  & \multicolumn{1}{c|}{96}    & 24                                  & 48                                  & 96                         &                          \\ \hline
\multicolumn{14}{c}{Discriminative   Score (Lower the better)}                                                                                                                                                                                                                                                                                                                                                                                                               \\ \hline
Original UNet & 0.711                      & 0.732                               & 0.686                               & 0.903                               & 0.956                               & 0.973                               & 0.713                      & 0.679                               & 0.752                      & 0.704                               & 0.830                               & 0.778                      & 0.785                    \\
TR-UNet       & \textbf{0.628}             & 0.778                               & 0.637                               & \textbf{0.772}                      & 0.876                               & 0.958                               & 0.640                      & \textbf{0.681}                      & 0.766                      & 0.690                               & 0.678                               & 0.763                      & 0.739                    \\
TD-UNet       & 0.704                      & \textbf{0.661}                      & 0.751                               & 0.842                               & 0.898                               & 0.765                               & 0.707                      & 0.831                               & \textbf{0.601}             & \textbf{0.619}                      & 0.898                               & \textbf{0.689}             & 0.747                    \\
Proposed      & 0.640                      & 0.897                               & \textbf{0.611}                      & 0.808                               & \textbf{0.664}                      & \textbf{0.755}                      & \textbf{0.706}             & 0.886                               & 0.672                      & 0.695                               & \textbf{0.659}                      & 0.722                      & \textbf{0.726}           \\ \hline
\multicolumn{14}{c}{Predictive Score (Lower the better)}                                                                                                                                                                                                                                                                                                                                                                                                                     \\ \hline
Original UNet & 18.311                     & 14.005                              & 15.527                              & 25.091                              & 26.966                              & 24.013                              & \textbf{20.273}            & 18.118                              & 20.995                     & 29.769                              & 28.881                              & 26.048                     & 22.333                   \\
TR-UNet       & \textbf{17.297}            & 13.717                              & 15.570                               & 26.860                               & 25.860                               & 24.882                              & 20.424                     & 17.382                              & \textbf{15.065}            & 29.913                              & 27.220                               & 25.796                     & 21.665                   \\
TD-UNet       & 17.397                     & 13.776                              & 15.213                              & 25.918                              & 25.596                              & 23.221                              & 20.311                     & 18.095                              & 15.835                     & 29.286                              & 25.992                              & \textbf{25.603}            & 21.354                   \\
Proposed      & \multicolumn{1}{l}{17.547} & \multicolumn{1}{l}{\textbf{13.616}} & \multicolumn{1}{l}{\textbf{14.472}} & \multicolumn{1}{l}{\textbf{24.039}} & \multicolumn{1}{l}{\textbf{24.408}} & \multicolumn{1}{l}{\textbf{23.152}} & \multicolumn{1}{l}{20.377} & \multicolumn{1}{l}{\textbf{17.184}} & \multicolumn{1}{l}{15.554} & \multicolumn{1}{l}{\textbf{29.247}} & \multicolumn{1}{l}{\textbf{27.072}} & \multicolumn{1}{l}{26.167} & \textbf{21.070}          \\ \hline
\end{tabular}
\end{table*}

\subsection{Comparison with the State-of-the-art Methods}

To verify the superiority of the proposed methods, some state-of-the-art models are trained on the industrial dataset as a comparison, including GAN-based approaches for time series (TTS-GAN\cite{li2022tts}, TimeGAN\cite{yoon2019time}, GAN-LSTM\cite{lu2021deep}) and diffusion model-based approaches for time series and tabular (DiffWave\cite{kong2020DiffWave}, SSSD\cite{lopezalcaraz2022diffusionbased}, Tabddpm\cite{kotelnikov2023tabddpm}). Among them, DiffWave\cite{kong2020DiffWave} is a diffusion model using a bidirectional dilated convolution architecture, which performs well in the industrial datasets of this experiment and is our main comparison method. To compare performances under different generation conditions, we vary the input length and the corresponding length of synthetic samples with a wide range: 24, 48, 96.

Table \ref{tab:comparation-CMAPSS} shows the experiment results of our method and other generative networks. Under the same length setting, Diff-MTS achieves consistent state-of-the-art performance in most datasets and length settings. 
For example, with the length setting of 96 in sub-dataset FD001, the Diff-MTS achieves a discriminative score of 0.611, which outperforms the DiffWave network by 32.4 \%. In the same length setting, Diff-MTS gives 33.4\% (21.744→14.472) predictive score reduction in FD001, 13.2\% (30.122→26.152) in FD002, 45.2\% (28.433 →15.554) in FD003, 28.7\% (32.515 →23.167) in FD004.
This signifies that Diff-MTS is able to provide more accurate generation results. Note that the GAN-based methods generally obtain a discriminative score above 0.94, which means synthetic time series obtained by these methods have large distributional differences between the real time series. This situation of GAN-based methods may result from their limitation for generating complex MTS data and unstable training, while Diff-MTS proposes Ada-MMD to enhances the condition consistency of the diffusion model and TDR-UNet to improve diffusion model’s ability to extract the temporal information, thus achieving superior results.

\textbf{\textit{FEMTO Dataset}}: We also perform an additional comparison experiment on the FEMTO dataset and vary the input length and the corresponding prediction lengths: 80, 160, 320. It can be observed that Diff-MTS consistently outperforms other methods at all length settings. The extensive experiment on different situations demonstrates the superior effectiveness of Diff-MTS in generating industrial multivariate time series.

\begin{table}[!tpb]
  \centering
  \caption{Ablation Study of the Conditional Ada-MMD Mechanism.}
  \begin{tabular}{cccc}
    \toprule
Model Name & \begin{tabular}[c]{@{}c@{}}Diff-MTS \\ w/o Ada-MMD\end{tabular} & \begin{tabular}[c]{@{}c@{}}Diff-MTS \\ w/o Ada\end{tabular} & \begin{tabular}[c]{@{}c@{}}Original\\ Diff-MTS\end{tabular} \\ 
    \midrule
FD001-48     & 14.667               & 13.976           & \textbf{13.616} \\
FD002-48     & 25.815               & 25.596           & \textbf{24.408} \\
FD003-48     & 17.950               & 18.350           & \textbf{17.184} \\
FD004-48     & 28.926               & 27.220           & \textbf{27.072} \\     
\bottomrule
  \end{tabular}
    \label{tab:ablation-mmd}
\end{table}

\subsection{Ablation Study of the Conditional Ada-MMD Mechanism}
The ablation experiment of the conditional adaptive MMD mechanism assessed its effectiveness through modifications to Ada-MMD module on four sub-datasets. We adjusted the Diff-MTS model and reported predictive scores for the generated results: (1) Diff-MTS w/o Ada-MMD represents the model with the exclusion of the Ada-MMD mechanism, (2) Diff-MTS w/o Ada indicates the model a model using pure MMD, and (3) the original Diff-MTS architecture.

As shown in Table \ref{tab:ablation-mmd}, the removal of the Ada-MMD mechanism in Diff-MTS struggling to capture complex distributions in the data, consequently leading to relatively lower predictive and discriminative scores. Specifically, Diff-MTS w/o Ada-MMD achieves an average predictive score of 21.840, representing a decrease compared to the original Diff-MTS. Additionally, the exclusion of the adaptive kernel in Ada-MMD also demonstrates a declining trend. In conclusion, these experiments highlight the significance of the Ada-MMD mechanism and the effectiveness of the adaptive kernel in the Ada-MMD mechanism.

\subsection{Ablation Study of the TDR-UNet}

TDR-UNet is characterized by time series decomposition and convolutional attention mechanism. To analyze the effectiveness of key components, we modified Diff-MTS and reported the predictive and discriminative scores: (1) TR-UNet means the model without the time series decomposition, (2) TD-UNet means the model without the convolutional attention mechanism, and (3) the original 1-D UNet architecture.

In Table \ref{tab:ablation-att}, one can observe that both the two elements significantly contribute to improving the quality of the generated time series data. The deployment of decomposition reconstruction UNet makes the network capture complex temporal patterns, thus improving the generation quality. Specifically, the TD-UNet achieves a discriminative score of 0.747 and a predictive score of 21.354 on average, which outperforms the original UNet by 4.8\% (0.785→0.747) and 4.4\% (22.333→21.354). It also can be observed that the TR-UNet outperforms the original UNet on average, which can benefit from focusing on the important regions in the reconstruction process.
In addition, we find that the combination of the decomposition module and reconstruction module improves generative performance on average. Overall, the ablation experiment demonstrates that both elements of the temporal decomposition reconstruction contribute to the quality of the synthetic sensor time series.

\begin{table}[htpb]
\centering
\caption{Comparison Result with other Methods Of Distance on C-MAPSS Datasets}
\label{tab:distance} 
\resizebox{\linewidth}{!}{\begin{tabular}{lllllllll}
\hline
\multicolumn{1}{c}{Method}   & \multicolumn{2}{c}{TTS-GAN\cite{li2022tts}}                      & \multicolumn{2}{c}{GAN-LSTM\cite{lu2021deep}}                     & \multicolumn{2}{c}{DiffWave\cite{kong2020DiffWave}}                     & \multicolumn{2}{c}{Diff-MTS}                     \\ \hline
\multicolumn{1}{c}{Distance} & \multicolumn{1}{c}{DTW} & \multicolumn{1}{c}{FD} & \multicolumn{1}{c}{DTW} & \multicolumn{1}{c}{FD} & \multicolumn{1}{c}{DTW} & \multicolumn{1}{c}{FD} & \multicolumn{1}{c}{DTW} & \multicolumn{1}{c}{FD} \\ \hline
\multicolumn{1}{c}{FD001-24} & 476.151                 & 170.298                & 2.948                   & 0.742                  & 2.367                   & 0.569                  & 2.272                   & 0.547                  \\
\multicolumn{1}{c}{FD001-48} & 499.358                 & 120.350                & 9.859                   & 1.476                  & 3.236                   & 0.568                  & 3.189                   & 0.564                  \\
FD001-96                     & 391.928                 & 83.847                 & 14.463                  & 1.603                  & 4.451                   & 0.595                  & 4.273                   & 0.563                  \\
FD002-24                     & 470.910                 & 117.733                & 18.033                  & 4.406                  & 4.975                   & 2.059                  & 4.986                   & 2.062                  \\
FD002-48                     & 533.100                 & 91.857                 & 15.347                  & 2.511                  & 6.480                   & 2.037                  & 6.506                   & 2.045                  \\
FD002-96                     & 539.056                 & 92.340                 & 13.942                  & 2.713                  & 8.745                   & 2.051                  & 8.714                   & 2.022                  \\
FD003-24                     & 353.075                 & 77.079                 & 2.492                   & 0.625                  & 2.285                   & 0.581                  & 2.379                   & 0.594                  \\
FD003-48                     & 508.226                 & 131.394                & 9.629                   & 1.446                  & 8.156                   & 1.602                  & 3.266                   & 0.614                  \\
FD003-96                     & 506.856                 & 130.327                & 10.783                  & 1.212                  & 4.609                   & 0.752                  & 4.377                   & 0.674                  \\
FD004-24                     & 478.746                 & 115.877                & 6.867                   & 2.386                  & 4.910                   & 2.044                  & 4.909                   & 2.032                  \\
FD004-48                     & 476.554                 & 115.427                & 8.076                   & 2.049                  & 6.445                   & 2.038                  & 6.458                   & 2.034                  \\
FD004-96                     & 577.966                 & 82.875                 & 9.447                   & 2.503                  & 8.645                   & 2.048                  & 8.676                   & 2.048                  \\
Average                      & 484.327                 & 110.784                & 10.157                  & 1.973                  & 5.442                   & 1.412                  & \textbf{5.000}          & \textbf{1.317}         \\ \hline
\end{tabular}
}
\end{table}

\subsection{Analysis on the Distance}
To quantify the distance between synthetic data and real time series, we use two distance functions (similarity measures): Dynamic Time Warping (DTW) distance and Frechet Distance (FD) functions. Table \ref{tab:distance} shows the DTW and FD distance of our methods and some baselines (DiffWave\cite{kong2020DiffWave}, GAN-LSTM\cite{lu2021deep}, TTS-GAN\cite{li2022tts}). Each method generates the same number of time series with the same labels (health indicator) as the original dataset.

From Table \ref{tab:distance}, it can be observed that the diffusion model-based methods (Diff-MTS, DiffWave) consistently outperform the GAN-based methods (TTS-GAN, GAN-LSTM). This is consistent with the results reported in Table \ref{tab:comparation-CMAPSS}, indicating that the diffusion method produces high-quality time series. In some settings, DiffWave slightly outperforms Diff-MTS, such as FD001 with lengths of 96. This could be attributed to certain unique characteristics of the datasets or specific noise factors in some  scenarios. However, it is essential to note that, on average, Diff-MTS still outperforms DiffWave on both similarity measures, indicating its ability to generate synthetic data that closely resembles real time series in a variety of scenarios.

\subsection{Visualization Analysis}
We utilize PCA and t-SNE techniques to reduce the synthetic and real data to two dimensions and plot their scatterplots in Fig. \ref{fig:prediction plot1}. In the scatterplots, synthetic data are denoted by brown dots while real data are represented by blue dots. A greater degree of overlap between these dots indicates higher similarity between the two data types. In the first column, the scatterplots of Diff-MTS demonstrate a notably improved overlap, especially when compared to other GAN-based benchmarks, suggesting that the synthetic datasets produced by diffusion model-based methods closely resemble the original data. For the PCA visualization of DiffWave, it can be observed that the blue and brown dots on the right side of the figure do not overlap well. It means that there are some distribution differences between the synthetic data of DiffWave and the real data.

Moreover, we display some synthetic data and real data to show our classifier-free guidance in Fig. \ref{fig:curves}. The black curves are real time series, while t The gold, blue and green curves represent the synthesized time series of Diff-MTS, DiffWave and TimeGAN, respectively. It can be observed that Diff-MTS can generate high-fidelity time series in the given condition and can generate more similar time series than DiffWave and TimeGAN. For example, the synthetic signal of TimeGAN in sensor s13 is only in the upper half of the diagram. In contrast, the synthetic signal of the Diff-MTS in sensor S13 represents the trend of the real time series.
 
\section{Conclusion}
In this paper, we propose Diff-MTS, a temporal-augmented conditional adaptive diffusion model for synthesizing industrial multivariate time series (MTS) data. Diff-MTS effectively mitigates the issues of non-convergence and unstable training encountered in Generative Adversarial Networks (GANs) and is capable of recovering high-fidelity signals from Gaussian noise. Comprehensive experiments demonstrate that Diff-MTS outperforms GAN-based methods, offering promising solutions for training industrial large models by generating high-quality industrial data.

In future work, we plan to investigate the integration of large language models (LLM) with industrial signal generation models. For example, by combining LLM with a time series prediction model, LLM can enhance its feature representation capabilities. In addition, given MTS data for industrial equipment, LLM can provide information about equipment failure states or health indicators and provide maintenance recommendations.

\bibliographystyle{ieeetr}
\bibliography{reference}

\begin{IEEEbiography}[{\includegraphics[width=1in,height=2.5in,clip,keepaspectratio]{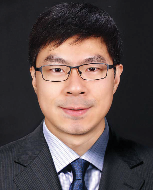}}]{Lei Ren} (Member, IEEE) received the Ph.D. degree in computer science from the Institute of Software, Chinese Academy of Sciences, Beijing, China, in 2009. 
	
He is currently a Professor with the School of Automation Science and Electrical Engineering, Beihang University, Beijing, China, and also with Zhongguancun Laboratory, Beijing, China. His research interests include neural networks and deep learning, time series analysis, and industrial AI applications. Dr. Ren serves as an Associate Editor for the IEEE Transactions on Neural Networks and Learning Systems and other international journals.	
\end{IEEEbiography}

\begin{IEEEbiography}[{\includegraphics[width=1in,height=2.5in,clip,keepaspectratio]{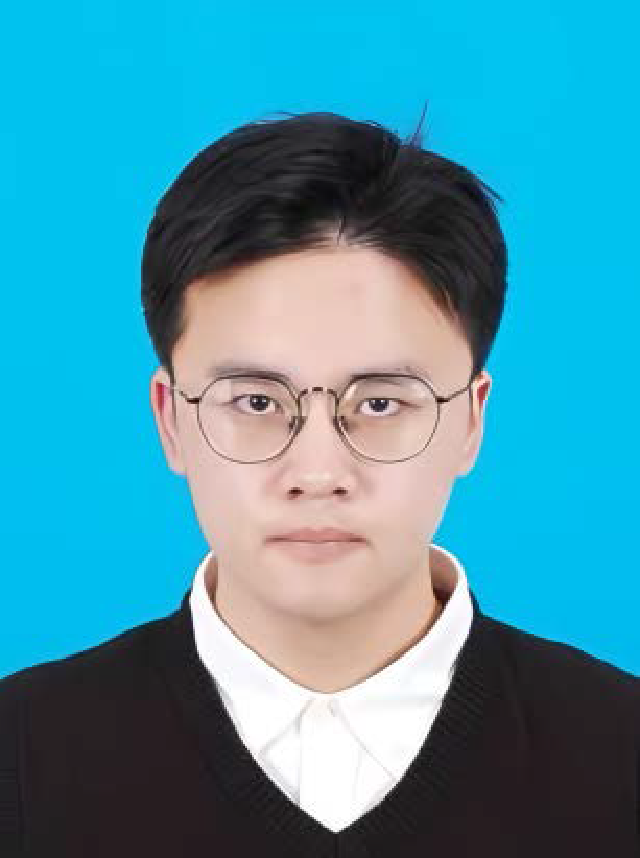}}]{Haiteng Wang} (Student Member, IEEE) received the B.Eng. Degree in automation engineering from Beihang University, Beijing, China, in 2021, where he is currently pursuing the Ph.D. degree with the School of Automation Science and Electrical Engineering. His current research interests include Time Series Prediction, Generative Models, and Edge Computing.
\end{IEEEbiography}

\begin{IEEEbiography}[{\includegraphics[width=1in,height=3in,clip,keepaspectratio]{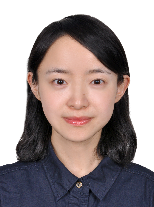}}]{Yuanjun Laili} (Member, IEEE) 
is an Associate Professor with the School of Automation Science and Electrical Engineering, Beihang University, Beijing, and with Zhongguancun Laboratory, Beijing. She received the B.S., M.S., and Ph.D. Degree from the School of Automation Science and Electrical Engineering at Beihang University. She is a chief scientist of the National Key R$\&$D Program of China and a member of IEEE and SCS (The Society For Modeling and Simulation International).

She has won the ”Young Talent Lift Project” supported by China Association for Science and Technology and the ”Young Simulation Scientist Award” from SCS. She serves as an Associate Editor of ”International Journal of Modeling, Simulation, and Scientific Computing” and ”Cogent Engineering”. Her main research interests are in the areas of intelligent optimization, modeling and simulation of manufacturing systems.
\end{IEEEbiography}

\end{document}